\documentclass[journal]{IEEEtran}
\ifCLASSINFOpdf
\else
\fi

\usepackage{amsmath,amsthm,verbatim,amssymb,amsfonts,amscd, graphicx}
\usepackage{graphics}
\usepackage{tabu}
\newtheorem{Proposition}{Proposition}
\newtheorem{theorem}{Theorem}
\long\def\symbolfootnote[#1]#2{\begingroup%
	\def\thefootnote{\fnsymbol{footnote}}\footnote[#1]{#2}\endgroup} 

\begin{document}
%
\title{Feature Selection Based on Wasserstein Distance}
\author{\IEEEauthorblockN{Fuwei Li}}
\maketitle
\symbolfootnote[0]{
This work was accomplished at UC, Davis, during Fuwei Li's (e-mail: fuweily@gmail.com) Ph.D. study under the supervision of Lifeng Lai and Shuguang Cui.
}

\begin{abstract}
This paper presents a novel feature selection method leveraging the Wasserstein distance to improve feature selection in machine learning. Unlike traditional methods based on correlation or Kullback-Leibler (KL) divergence, our approach uses the Wasserstein distance to assess feature similarity, inherently capturing class relationships and making it robust to noisy labels. We introduce a Markov blanket-based feature selection algorithm and demonstrate its effectiveness. Our analysis shows that the Wasserstein distance-based feature selection method effectively reduces the impact of noisy labels without relying on specific noise models. We provide a lower bound on its effectiveness, which remains meaningful even in the presence of noise. Experimental results across multiple datasets demonstrate that our approach consistently outperforms traditional methods, particularly in noisy settings.
\end{abstract}


%
\IEEEpeerreviewmaketitle

\section{Introduction}
%
%
%
%

Feature selection is a thoroughly studied topic in recent decades. In the preliminary phase of this research topic, it is aimed to increase the performance of the machine learning algorithm by selecting the most important features.  At that time, machine learning algorithms are far less powerful than today's, so by reducing the dimension of the input data features, the algorithm can learn much faster and more accurate. Also, the limited storage capacity at that time required the data samples to be as small as possible while keeping the most informative features. Now, we have much more powerful and efficient machine learning algorithms. For example, the deep neuron networks can handle billions of data samples. It seems that there is no need to do feature selection anymore. However, the data generated by humans and Internet of things(IoTs) have also increased dramatically. The rate grows even faster than the storage capacity as well as the computing power. In the near future when the IoTs play a more important part in our life, the data generated by IoTs and humans will be expected to grow exponentially.

Even though we have larger storage capacity and greater computing power, there are still a lot of potential benefits to doing feature selection. First, feature selection can speed up the computation of machine learning. There are two phases in today's machine learning algorithm, the training phase and the test phase. In the training phase, the parameters of the learning machine grow with the dimension of the input data.
For example, the deep neuron network, facilitated by hundreds of network layers and thousands of layer width, while making the training and inference very slow. If the dimension of the input feature is reduced, it will make learning machines more efficient and speed up the training and inference process. Second, it can help us get better generalization performance through reducing the dimension of the input data. It is proved that with fixed number of samples, the more powerful of the learning machine, the more likely it will overfit the data and thus poorer generalization performance.
Even the data dimension is high, only small parts of it contains the most useful information \cite{fan2014challenges}. 
By feature selection, we can distinguish those important features among the high dimensional features of the data, use simpler learning machines, and get better generalizations. Third, it can also reduce the storage capacity to low the cost. 
Though the price of the storage device reduce almost linearly over years\cite{websitestorage}, the data grows exponentially. If the dimension of the data is reduced, it will be economically efficient. 

Algorithms of feature selection can be divided into three categories\cite{guyon_introduction_2003}: wrapper, embed, and filter methods. 
Wrapper methods use some exhaustive searching among different subsets of features and evaluate the feature sets based on a specific learning machine.
These methods lead to very high computational complexity and is feasible only when the number of the features is small, it is seldom used nowadays.

The embed methods integrate feature selection into the design of the learning machine. Therefore, feature selection is accomplished since the training of the learning machine. The embed method includes various kinds of sparse driven algorithms\cite{tibshirani1996regression}\cite{chen2006combining}. Since feature selection is done together with the training, they are time-saving. However, these methods require careful designing. Different learning algorithms and different parameters may lead to totally different features. 

The filter method is trying to design a scheme independent of the learning machines. 
Hence, compared with the wrapper and embed methods, it is suitable for broader kinds of learning machines. 
So, in this paper, we will focus on the filter method. 


Within filter methods, various kinds of criteria are proposed to select the best feature set, including the correlation based criterion\cite{hall1999correlation}, 
probability measure based criterion\cite{koller1996toward}, and conditional likelihood based criterion\cite{brown_conditional_2012}.
Since most of the machine learning algorithms are intend to use a parametric model to mimic the empirical data distribution under some probabilistic distance measure,
it is natural to choose the probability measure as the feature selection criterion. 
Traditionally, people may use measures induced by information theory,
for example, the Kullback-Leibler (KL) divergence, f-divergence, and $\chi^2$ divergence. Among the three, the KL divergence is widely used\cite{koller1996toward}\cite{brown_conditional_2012}\cite{peng_feature_2005}. Even though KL divergence is simple, it still has some weaknesses. The definition of KL divergence between $p(Y|\mathbf x)$ and $p(Y|\mathbf x_\theta)$ is $\sum_{i=1}^{n_c} p(y=i|\mathbf x)\ln\frac{p(y=i|\mathbf x)}{p(y=i|\mathbf x_\theta)}$, where $n_c$ is the number of classes, $Y$ is label of the class, $\mathbf x$ is the whole feature set before feature selection, and $\mathbf x_\theta$ is a subset of feature indexed by $\theta$. 
The definition indicates the KL divergence treats all the classes equally. For example, if $p(Y|\mathbf x) = [0.5,0.5]$ and $p(Y|\mathbf x_{\theta_1})=[0.1,0.9]$, and another distribution is $p(Y|\mathbf x_{\theta_2})=[0.9,0.1]$, then we have 
$KL[p(Y|\mathbf x), p(Y|\mathbf x_{\theta_1})] = KL[p(Y|\mathbf x), p(Y|\mathbf x_{\theta_2})]$.
Even if we have two totally different distributions $p(Y|\mathbf x_{\theta_1})$ and $p(Y|\mathbf x_{\theta_2})$, the KL distance measure can not distinguish them. 

\begin{figure}[t]
\centering
\includegraphics[width=.7\linewidth]{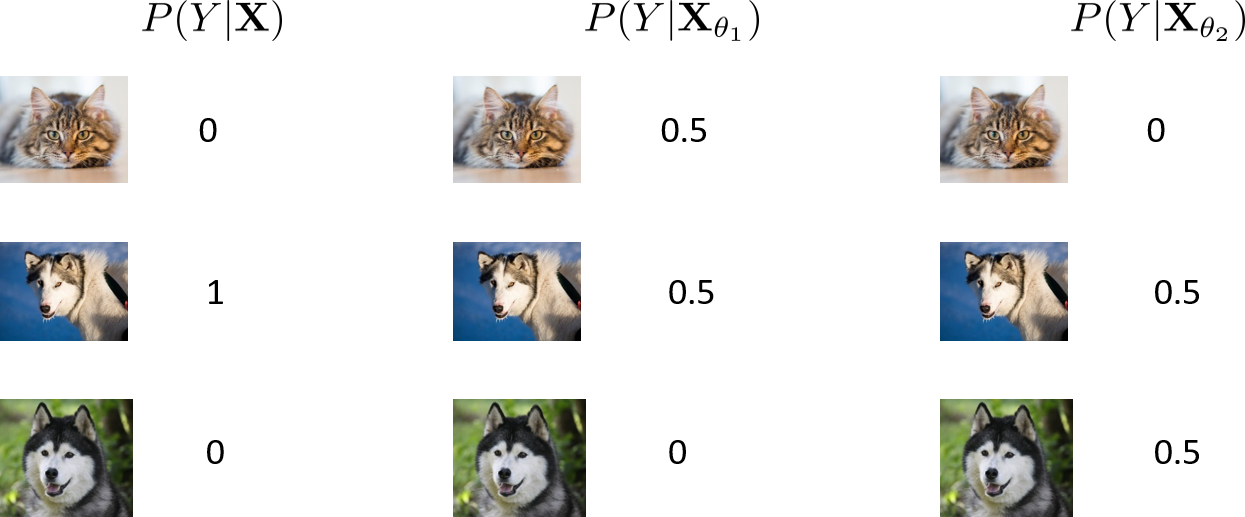}
\caption{The first column indicates the original conditional distribution for different classes. The second column indicates the conditional distribution based on feature set $\theta_1$. The third column shows the conditional distribution based on feature subset $\theta_2$. If we use KL distance as the features selection criterion, the two selected feature subset will lead to the same distance. So, there will be no differences if we select those two feature sets. However, intuitively, the second feature subset will give us more reasonable results because the husky is more similar to the alaska than the cat.}
\label{intuitive}
\end{figure}

In the computation of KL distance, the denominator in the log term also makes the KL divergence unstable. Consequently, when the denominator is approximate to zero, the KL divergence will explode. This property makes feature selection based on the KL distance not stable especially when some noise is introduced into the data samples. 

One reason why KL distance is so widely used is
its simple and decomposable computation. 
With this decomposable computing advantage, we can design efficient machine learning algorithms. However, it fails to capture the relationship between different classes. Relationships between classes are very common in real practice. For example, in the text classification problem, one article is assigned to several classes which have hierarchical structures\cite{webtextclass}. In ImageNet\cite{russakovsky2015imagenet}, some of the image labels have very high correlation. Several papers have used this correlation in classification problems\cite{frogner_learning_2015}\cite{bi_multilabel_2014}\cite{mccallum1998improving}. Nevertheless, little work has done in the context of feature selection. In this work, we assume there is a ground metric, \emph{a prior}, describing the dissimilarities between classes.
Fig.\ref{intuitive} gives a simple example to show how useful of this inter-class information is. Suppose there are three classes: cat, husky and alaska. 
If the distribution based on the full features is $p(Y|\mathbf x) =[0,1,0]$ and two distributions based on different sub features $p(Y|\mathbf x_{\theta_1})=[1,0,0]$ and $p(Y|\mathbf x_{\theta_2})=[0,0,1]$. In this case, we would better choose $\mathbf x_{\theta_2}$ as the selected feature due to the fact that confusing a dog with a cat is more severe than confusing with two kinds of dogs. However, if we ignore the class similarity and use KL divergence, it can not tell which subset of feature is better. While, Wasserstein distance is a natural probabilistic distance measure which inherently use the class similarities. 

Wasserstein distance is also called Earth Movement distance\cite{villani_optimal_2009}.
Wasserstein distance is used in many applications including computation graphics\cite{gangbo2000shape}, clustering\cite{applegate2011unsupervised}, Generative Adversarial Networks\cite{arjovsky_wasserstein_2017}, and so forth. In the latter section, we will detail the computation of Wasserstein distance and how it is used in our feature selection problem. 

Another common problem many machine learning tasks encounter is the label noise. It is known that the label noise will dramatically decrease the classifiers' performance\cite{zhu_class_2004}. Sources of label noise may come from various ways. First, label noise may simply come from the transmission process. For example, if the class labeling work is distributed to the workers by the wireless communication channel, the communication noise may add to the labels\cite{zhu_class_2004}. Second, label noise may come from the lack of information. For instance, 
to tell whether a person is male or female,
providing his/her voice data is not sufficient. In medical diagnosis, a doctor usually uses several reports provided by different inspection methods to determine a person has a certain kind of illness. Third, labels may come from some unreliable sources.  In particular, now we have billions of data need to be labeled and the price to do label by human experts is very high. Instead, people are trying to use some cheap but not reliable source, like Amazon Mechanical Turk\cite{ipeirotis2010quality}\cite{raykar_learning_2010}. Fourth, when the class is subjective, for example, in medical diagnosis, for a certain medical image, different experts may give different results. This happens especially when two classes have some conceptual confusions. So, label noise commonly exists in real practical dataset. 

Various methods have been introduced to tackle the label noise problem\cite{frenay_classification_2014}. For example, the filter based methods firstly detect the wrongly labeled data, correct them, and then use a traditional algorithm to do classification. Others use a specific label noise model to capture how label noise is introduced\cite{sukhbaatar_learning_2014} \cite{xiao2015learning} \cite{natarajan_learning_2013}. Even though so many label noise models are proposed, different models can only capture one certain aspect of the label noise. There is no unified way to model the label noise. In real practice, it is even more complicated, because in most cases, there is no ground truth label noise model. Though there is no unified way to describe the label noise, all of them admit that label noise is more likely to happen between two similar objects. We will show that our Wasserstein distance based algorithm captures this property naturally. Thus our algorithm is more robust to the label noise. 


The contribution of this work can summarized as follows:
\begin{itemize}
    \item 
    We are the first to use Wasserstein distance in the context of feature selection. In the existing works, people tended to use information based criteria to choose the most important features. Among which, KL divergence is the most widely used criterion. However, it lacks the ability to capture the correlation between classes. Wasserstein distance has the inherent nature to capture the similarities between classes. Thus makes it more suitable to more commonly seen feature selection problem. To show the Wasserstein distance is suitable for feature selection problem, we prove several useful Wasserstein distance induce independence properties. With the facility of the Wasserstein distance, we proposed a Markov blanked based feature selection algorithm and analyzed the computational complexity of the algorithm. 
    \item
    We analyzed the performance under the noisy labeled settings and showed that Wasserstein distance based feature selection has the potential to reduce the impact of noise labels.  Noisy labels exists in many practical feature selection problems and how the noisy labels were introduced is every complex. Directly model the noisy label is impractical. Our analysis does not depend on a specific noisy model. With the facility of the Wasserstein distance, we give a lower bound on how good it is using the Wasserstein distance to do feature selection even some noisy labels exist. Also, we show that in general this bound is small and meaningful under the Wasserstein distance. 
    \item
    We do several simulations, which includes text analysis, email classification, and image classification, to show our proposed Wasserstein distance based feature selection algorithm is useful and superior to its corresponding KL distance based feature selection algorithm. 
\end{itemize}

The outline of the subsequent parts is: we formulate the feature selection problem rigorously, introduce the Wasserstein distance, give a brief introduction to Markov Blanket, show the independence properties of Wasserstein distance, and introduce our algorithm; analyze the performance of our algorithm in the noisy label settings under the Wasserstein distance; demonstrate the performances under serveral simulations. 

\section{The Algorithm}

\subsection{Problem Formulation}
Firstly, it is necessary to clarify the feature selection problem. Suppose there are a bunch of data samples $\{\mathbf x_i,y_i\}_{i=1}^N$, where $N$ is the number of samples, $\mathbf x_i$ is the feature vector for sample $i$, and $y_i$ is the label for sample $i$. $\mathbf x_i \in \mathbb{R}^M$ is a $M$ dimensional vector and $\mathbf x_i =[x_1^i, x_2^i, \cdots, x_M^i]$. Each component of $\mathbf x_i$, $x_j^i$, is called a feature of samples $i$. For single labeled data, the label $y_i$ belongs to one of the class labels $y_i \in \{c_1,c_2,\cdots,c_{n_c}\}$ and for multi labeled data, the label $y_i$ belongs to subset of the class labels $y_i \subseteq \{c_1,c_2,\cdots,c_{n_c}\}$, where $n_c$ denotes the number of classes. Further, all the samples are assumed to be independent and identically distributed according to distribution $p(\mathbf X, Y)$. We also use the set $\theta \subseteq \{1,2,\cdots,M\}$ to represent the index of the feature we are choosing and $\mathbf X_\theta$ denotes the selected features which are indexed by $\theta$. 

For a learning problem, given an instance of the feature $\mathbf X=\mathbf x$, we want to know the distribution of the classes. This class conditional distribution can further used to construct an optimal \emph{Bayesian estimator}.
So, the class conditional distribution is most important in a learning problem. 
After feature selection, we are only given a subset of the feature, $\mathbf X_\theta  = \mathbf x_\theta $, we still intend to make the class distribution be similar to the original class distribution given the whole features. 
So, we minimize  the distance between the two conditional distributions, the class conditional distribution given all the features $p(Y|\mathbf X=\mathbf x)$ and the class conditional distribution given the selected features $p(Y|\mathbf X_\theta=\mathbf x_\theta)$. Formally, we attempt to solve this minimization problem
\begin{align*}
\underset{|\theta|=K}{\text{argmin}}
\, M \,[\,p(Y|\mathbf x),\, p(Y|\mathbf x_\theta)\,], 
\end{align*}
where $M\,[\cdot, \cdot]$ is some probabilistic distance measure to tell the distance between two probability distributions, $|\theta|$ means the cardinality of the set $\theta$ and $K$ is the number of features we want to choose. Given the weights of different instances of feature $\mathbf X=\mathbf x$, it is better to use the expected value of $M$ as the distance measure. So, our minimization problem is 
\begin{align}
\underset{|\theta|=K}{\text{argmin}}
\, \mathbb{E}_{\mathbf X} M\,[\,p(Y|\mathbf X),\, p(Y|\mathbf X_\theta)\,],
\label{origi_opt}
\end{align}
In this paper, we choose $M$ as the Wasserstein distance. We will explain the details of the Wasserstein distance in the following. 

\subsection{The Wasserstein Distance}

Wasserstein distance is also called earth mover's distance\cite{rubner2000earth} or optimal earth transport distance\cite{villani_optimal_2009}. Suppose there are two measures $\nu_1$ and $\nu_2$ on $\mathcal{C}$, and a cost function $d: \mathcal{C}\times \mathcal{C} \to \mathbb{R}$ measures the cost of transporting the measure of a mass on $\mathcal{C}$ to another $\mathcal{C}$. The Wasserstein distance is defined as the cheapest way to transport the mass with the probability measure $\nu_1$ to the corresponding mass with the probability measure $\nu_2$:
\begin{align}
W(\nu_1,\nu_2) = \inf_{\pi \in \Pi(\nu_1,\nu_2) } \int_{\mathcal{C}_1\times \mathcal{C}_2} d(c_1,c_2) \pi(\mathrm{d}c_1, \mathrm{d}c_2)\label{original_wass},
\end{align}
where $\Pi(\nu_1,\nu_2)$ is the set of joint probability measure with marginal probability measure $\nu_1$ and $\nu_2$ respectively.
Solving this optimization problem means to find the joint probability which minimizes the cost of transporting one mass with measure $\nu_1$ to the corresponding mass with measure $\nu_2$. In this paper, we only focus on the discrete measure on the classes. 
In this case, the Wasserstein distance between probability measure $\mathbf p(Y)=[p(y_1),p(y_2),\cdots,p(y_{n_c})]$ and $\mathbf q(Y)=[q(y_1),q(y_2),\cdots,q(y_c)]$ can be computed by solving the linear optimization problem:
\begin{align}\nonumber
\min \quad &\text{tr}(\mathbf Q^\top \mathbf D) \\ \nonumber
\text{s.t.} \quad & \mathbf Q \cdot \mathbf 1 = \mathbf p(Y) \\ \nonumber 
&\mathbf Q^\top \mathbf 1 = \mathbf q(Y) \\ 
& Q_{i,j}\ge 0,\, \forall \, i,j \in \{1,2,\cdots,n_c\}, 
\label{wassopt}
\end{align}
where $\mathbf Q$ is the joint probability matrix with marginal distributions be $\mathbf p(Y)=[p(Y=c_1),p(Y=c_2),\cdots,p(Y=c_n)]$ and $\mathbf q(Y)=[q(Y=c_1),q(Y=c_2),\cdots,q(Y=c_n)]$; $\mathbf D$ is the distance matrix with its elements defined as $D_{ij} = d(c_i,c_j)$ and $d(c_i,c_j)$ defines the distance from class $c_i$ to class $c_j$ which measures the dissimilarity between class $c_i$ and class $c_j$. 
This is a linear programming problem. To solve this linear programming, the computational complexity would be $\mathcal O(n_c^{3.5})$\cite{nesterov1994interior}. The computation is heavy when the number of classes is large. Recently, \cite{cuturi_sinkhorn_2013} proposed a regularized form of this problem. They added a negative entropy of the joint distribution matrix regularization term in the objective function. Accordingly, we are solving the minimization problem with an objective function as 
$$\min: \text{tr}(\mathbf Q^\top \mathbf D)-\frac{1}{\lambda} H(\mathbf Q),$$
where $H(\mathbf Q)$ computes the entropy of the joint matrix $\mathbf Q$ and is defined as $H(\mathbf Q) = -\sum_{i,j}Q_{ij}\ln Q{ij}$. In this way, the new problem can be solved efficiently by a matrix balance iteration process. This entropy regularized algorithm results in a computational complexity of $\mathcal{O}(n_c^2)$ and maintains high accuracy. In our algorithm, we will use this method to compute the Wasserstein distance.

\subsection{Markov Blanket}

\begin{figure}[t]
\centering
\includegraphics[width=.4\linewidth]{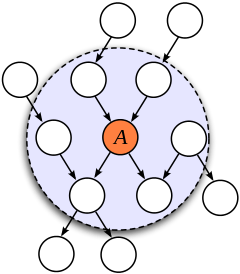}
\caption{The probabilistic graphic model, the Markov Blanket of node A contains the nodes in the shaded area.}
\label{MarkovBlanket}
\end{figure}

Markov Blanket is widely used in feature selection\cite{koller_toward_1996}\cite{margaritis2000bayesian}\cite{yaramakala2005speculative}\cite{pena2007towards}.
Markov Blanket is the smallest feature set which contains all the information about the target.
In a directed probabilistic graphic model, the Markov Blanket of a node is the nodes' parents, children and spouses. Fig.\ref{MarkovBlanket} shows a probabilistic graphic model, where the Markov Blanket of node A contains the nodes in the shaded area. More formally, the Markov Blanket of the target $Y$ is the smallest set $\mathcal{B}\subset \{1,2,\cdots,M\}$ which have $p(Y=y|\mathbf X=\mathbf x) = p(Y=y|\mathbf X_{\mathcal{B}}=\mathbf x_{\mathcal{B}})$ for every instance of $\mathbf X$.

Ideally, if we want to do feature selection without information loss, the Markov Blanket of the target is the best feature set we can get. 
Most of the features selection algorithms based on Markov Blanket mentioned above utilize the KL divergence distance. In this paper, we are going to introduce the Wasserstein distance into the feature selection. So, we will start from the basic feature selection algorithm described in \cite{koller_toward_1996}.
Then, we will show the benefits of using the Wasserstein distance in feature selection. 

To further explain our algorithm, we need elaborate some independence properties under the Wasserstein distance measure. 

\begin{Proposition}
$p(Y|\mathbf X_{\mathcal{G}_i},X_i) = p(Y|\mathbf X_{\mathcal{G}_i})$ iif $W [p(Y|\mathbf X_{\mathcal{G}_i},X_i),p(Y|\mathbf X_{\mathcal{G}_i})] = 0 $ for every $\mathbf{X}_{\mathcal{G}_i}=\mathbf{x}_{\mathcal{G}_i}, X_i = x_i$,
where $\mathcal{G}_i$ is a subset of the features, $i \notin \mathcal{G}_i$, and $W[\mathbf p,\mathbf q]$ denotes the Wasserstein distance between distribution $\mathbf p$ and $\mathbf q$. 
\label{theorem:independence}
\end{Proposition}

\begin{proof}
Suppose the elements of the distance matrix $\mathbf D$ in Problem(\ref{wassopt}) are positive except the diagonal elements and let $\mathbf p(Y) = p(Y|\mathbf X_{\mathcal{G}_i}=\mathbf x_{\mathcal{G}_i},X_i = x_i),\mathbf q(Y)=p(Y|\mathbf X_{\mathcal{G}_i}=\mathbf x_{\mathcal{G}_i})$. The solution of the optimization problem(\ref{wassopt}) is zero if and only if the joint distribution matrix $Q_{i,j}=0$ for all $i\neq j$. Then we have $\mathbf Q \cdot \mathbf 1 = \mathbf Q^\top \mathbf 1$, so $p(Y|\mathbf X_{\mathcal{G}_i}=\mathbf x_{\mathcal{G}_i},X_i = x_i) = p(Y|\mathbf X_{\mathcal{G}_i}=\mathbf x_{\mathcal{G}_i})$. Also, if $p(Y|\mathbf X_{\mathcal{G}_i}=\mathbf x_{\mathcal{G}_i},X_i = x_i) = p(Y|\mathbf X_{\mathcal{G}_i}=\mathbf x_{\mathcal{G}_i})$. It is easy to see that the optimal solution of problem(\ref{wassopt}) is zero.
\end{proof}

In Proposition \ref{theorem:independence}, $p(Y|\mathbf X_{\mathcal{G}_i},X_i) = p(Y|\mathbf X_{\mathcal{G}_i})$
implies $p(Y,X_i|X_{\mathcal{G}_i})=p(Y|X_{\mathcal{G}_i})\cdot p(X_i|\mathbf{X}_{\mathcal{G}_i})$. It tells us that the Wasserstein distance can reveal the independence property in the probabilistic graphic model. 


In paper \cite{koller1996toward}, they tried to find a Markov Blanket for each feature $X_i$. However, finding the Markov Blanket for a feature is too hard in practice. Proposition 2 tells us a more practical way to find whether $X_i$ belongs to the Markov Blanket.

\begin{Proposition}
 Let $\mathbf X_{\mathcal{G}_i}$ denotes a subset of $\mathbf X\setminus X_i$ and contains all the features which $X_i$ directed to. If for every setting of $Y=y, \mathbf X_{\mathcal{G}_i}=\mathbf x_{\mathcal{G}_i}, X_i = x_i$ we have $p(Y=y|\mathbf X_{\mathcal{G}_i}=\mathbf x_{\mathcal{G}_i},X_i = x_i)=p(Y=y|\mathbf X_{\mathcal{G}_i})$. Then, $X_i$ does not belong to the Markov Blanket. 
\end{Proposition}

\begin{proof}
 Since $X_i$ and $Y$ are independent given $\mathbf{X_{\mathcal{G}_i}}$ and $\mathbf X_{\mathcal{G}_i}$ contains all the nodes which $X_i$ direct to, if $X_i$ belongs to the Markov Blanket, there should be a node which both $Y$ and $X_i$ directed to, which contradicts the assumption that $Y$ and $X_i$ are independent given $\mathbf X_{\mathcal{G}_i}$. 
\end{proof}

As \cite{koller1996toward} has pointed out, even if we can find the appropriate feature set $\mathbf X_{\mathcal{G}_i}$, in practice, the distance of the two conditional probabilities, $p(Y|\mathbf X_{\mathcal{G}_i}=\mathbf x_{\mathcal{G}_i},X_i = x_i)$ and $p(Y|\mathbf X_{\mathcal{G}_i}=\mathbf x_{\mathcal{G}_i})$ , will hardly be zero. This is due to the noise or the limited samples. To test whether $p(Y|\mathbf X_{\mathcal{G}_i},X_i) = p(Y|\mathbf X_{\mathcal{G}_i})$, we need compare every configuration of $\mathbf X_{\mathcal{G}_i}$ and $X_i$. To make it more convenient and consider the weights of $\mathbf X_{\mathcal{G}_i}$ and $X_i$ into consideration, we use the averaged distance to measure their discrepancy. The averaged distance is defined as:
\begin{equation}
\delta(X_i) = \mathbb E_{\mathbf X_{\mathcal{G}_i},X_i}W[p(Y|\mathbf{X}_{\mathcal{G}_i},X_i), p(Y|\mathbf X_{\mathcal{G}_i})]
\label{deltaxi}.
\end{equation}

So, if there is a $\mathcal{G}_i$ which has $p(Y|\mathbf X_{\mathcal{G}_i}=\mathbf x_{\mathcal{G}_i},X_i=x_i) = p(Y|\mathbf X_{\mathcal{G}_i}=\mathbf x_{\mathcal{G}_i})$ for every $\mathbf X_{\mathcal G_i}=\mathbf x_{\mathcal G_i}, X_i = x_i$, $\delta(X_i)$ should be zero or at least small in practice.

To find an appropriate $\mathcal G_i$ for feature $X_i$, we also use the approximation method used in \cite{koller1996toward}. We find the top-L most relevant features with $X_i$ as $\mathbf X_{\mathcal G_i}$. 
So, by choosing the most relevant $L$ features as $\mathbf X_{\mathcal G_i}$, probably $\mathbf X_{\mathcal G_i}$ will include all the nodes inbound to or outbound from $X_i$ . From the probabilistic graphic model, those nodes will cut all paths from $X_i$ to $Y$.
According to Proposition 2 and the definition of $\delta(X_i)$ in Eq.(\ref{deltaxi}), if $\delta(X_i) = 0$, $X_i$ does not belong to the Markov Blanket, we can safely delete feature $X_i$. 

\subsection{Feature Selection Algorithm}
With help of Proposition 1 and Proposition 2, here we give our algorithm. 

First, compute the absolute value of correlation $\rho_{i,j}=|\frac{Cov(X_i,X_j)}{std(X_i)std(X_j)}|$ for each pair of feature $X_i$ and $X_j$. Those values are used in the following step to estimate the distance between two feature nodes in a probabilistic graphic model. Let $\mathbf{\bar X} = \mathbf X$ to denote the current existing features.  

Second, For every $X_i \in \mathbf{\bar X}$, find the features with $L$ largest value of $\rho _{i,j}$ and define them as $\mathbf X_{\mathcal{G}_i}$. In this step, we are trying to find all the nodes which $X_i$ directed to. According to the "folk-theorem"\cite{koller1996toward}, the closer of two nodes, the influence tend to be stronger. By choosing $L$ largest correlated nodes, we believe it will contain all the node $X_i$ directed to. 

Third, Compute $\delta(X_i)$ for each $X_i\in \mathbf{\bar X}$. According to Proposition 1 and Proposition 2, if $\delta(X_i)$ is zero, $X_i$ does not belong to the Markov Blanket. 

Fourth, Find $X_i$ which has the smallest $\delta(X_i)$, delete $X_i$ from $\mathbf{\bar X}$, and let $\mathbf{\bar X} = \mathbf{\bar X}\setminus X_i$. Because of the noise or limited sample effect, $\delta(X_i)$ will hardly be zero even if it is not contained in the Markov Blanket. So, for each run, we choose the one with smallest $\delta$ value and delete it from the remaining feature set. 

We summarize our algorithm in Table \ref{tab:algorithm}. 
\begin{table}[t]
    \caption{summary of our algorithm}
    \centering
    \begin{tabu} to 0.45\textwidth {  X[l]  }
        \hline
        1. Compute the absolute value of correlation $\rho_{i,j}=|\frac{Cov(X_i,X_j)}{std(X_i)std(X_j)}|$ for each pair of feature $X_i$ and $X_j$. Let $\mathbf{\bar X} = \mathbf X$.  \\ \\
        2. For every $X_i \in \mathbf{\bar X}$, find the features with $L$ largest value of $\rho _{i,j}$ and define them as $\mathbf X_{\mathcal{G}_i}$. \\ \\ 
        3. Compute $\delta(X_i)$ for each $X_i\in \mathbf{\bar X}$. \\ \\ 
        4. Find $X_i$ which has the smallest $\delta(X_i)$, delete $X_i$ from $\mathbf{\bar X}$, and let $\mathbf{\bar X} = \mathbf{\bar X}\setminus X_i$. \\ \\ 
        5. Return to (1) until the size of the $\bar{\mathbf X}$ equal to $K$. \\
        \hline
    \end{tabu}
    \label{tab:algorithm}
    \end{table}




\subsection{Complexity Analysis}

Suppose there are $N$ samples and $n_c$ classes, the number of features is $M$, the number of conditioned features is $L$ and $K$ is the number of features we want to choose. Below we will analyze the computational complexity of the algorithm in each step. 
\begin{enumerate}
\item Compute the feature correlation matrix. For computing each pair of the feature correlation, the complexity is $N$. Totally we have $\frac{M(M-1)}{2}$ pairs of features. So, the computational complexity is $\mathcal{O}(NM^2)$;
\item Sort the correlation matrix. For each feature, we would find the most $L$ significant features. The complexity of partial sorting algorithm to find the most relevant $L$ features is $\mathcal{O}(M+L\log L)$. Then the overall complexity of computing the correlation and sorting is $\mathcal{O}(M^2(N+M+L\log L))$;

\item To eliminate each feature, we need to calculate the Wasserstein distance between two conditional distributions. 
The complexity of estimating the conditional probability of $p(Y|\mathbf X_{\mathcal G_i})$ varies with different estimation algorithms. Here we assume all the features are discrete valued. Suppose the features are binary valued, and then the complexity to compute the conditional distribution is $2^LN$. The second step is to compute the distance between those two conditional probabilities. Computing the distance involves solving a linear programming problem. The complexity of the linear programming is $\mathcal{O}(n_c^3)$. By using an approximate method in \cite{cuturi_sinkhorn_2013}, the complexity can be reduced to $\mathcal{O}(n_c^2)$. So, the complexity of calculating the Wasserstein distance of the conditional distribution is $\mathcal O(2^L N  n_c^2)$;

\item Suppose, we are in the $i$'th step, then there are $M-i$ features left. Thus, we will need $M-i$ times of computing the Wasserstein distance. 
Then the complexity of eliminating process is $\mathcal{O}((2M-K)K2^LNn_c^2)$. In each eliminating step, we assume re-computing all the $\delta(X_i)$ for each remaining feature $X_i$. 
Actually, the eliminated feature in the previous step only effect only a small part of the conditional distributions. Then the first term can be reduced to a constant. 
\end{enumerate}
In summary, the computational complexity of the algorithm is 
$$\mathcal{O}(M^2(N+M+L\log L)+(2M-K)K2^L N n_c^2).$$
When $K$ is small
the complexity is approximately to 
$$\mathcal{O}(M^3+M^2N+M^22^LN n_c^2).$$
When the number of samples is far larger than the number of features, the complexity of the algorithm will be dominated by the term $N M^2 2^L n_c^2$.

\section{Feature Selection with Noisy Labels}
\subsection{Feature Selection Under Noisy Labels}
In this section, we will analyze the performance of our feature selection algorithm based on Wasserstein distance under the noisy labeled data samples. In real practice, the observed samples always contain noisy labels and we can only do feature selection based on these noisy labeled data samples. 
However, we want to know how good the selected features are when compared with the features selected using the clean data samples.

Label noise exists in most of the data set. Especially, label noise happens between similar classes. 
Suppose the original data set samples are $\{\mathbf x_i,y_i\}_{i=1}^N$ and they are drawn independent and identically from the joint distribution $p(\mathbf X, Y)$. 
However, because of the label noise what we observed is the noisy labeled data samples,  $\{\mathbf x_i,\tilde y_i\}_{i=1}^N$, and we assume they are drawn independent and identically from the joint distribution $p(\mathbf X, \tilde Y)$. Here, we assume only the labels of the data samples differ from  the original data samples. So, the marginal distributions of features are the same.
Since we can only access to the noisy labeled data samples, to select the optimal features, instead of solving the optimization problem Prob.(\ref{origi_opt}), we can only select the optimal features by solving this optimization problem 
\begin{equation}
\underset{|\theta|=K}{\text{argmin}}\,\mathbb E_{\mathbf X} W[p(\tilde Y|\mathbf X),\,p(\tilde Y|\mathbf X_{\theta})].
\label{noise_opt}
\end{equation}

\begin{figure}[t]
\centering
\includegraphics[width=.7\linewidth]{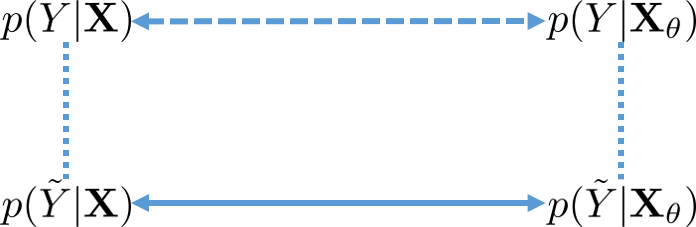}
\caption{The relationships between conditional distributions before feature selection and after feature selection in the original and observed data set respectively}
\label{cond_rela}
\end{figure}

Here, unlike most of the analysis, we do not assume a specific label noisy model\cite{frenay_estimating_2014} in our analysis. We still use the set $\theta \subseteq \{1,2,\cdots,M\}$ to denote the selected feature set. 
Fig.(\ref{cond_rela}) shows the relationships between the conditional distributions before and after feature selection in the original clean data set as well as in the observed noisy labeled data set. 

Before further analysis, we need define some quantities to simplify the notation of our analysis. 
First, 
$\epsilon_1$ refers to
the expected Wasserstein distance between the original conditional distribution and the noisy labeled conditional distribution, 
$$\epsilon_1 = \mathbb E_{\mathbf X}W[p(Y|\mathbf X),\, p(\tilde Y|\mathbf X)].$$

Second, 
$d_1(\theta)$ refers to 
the expected Wasserstein distance between the conditional distribution before feature selection and after feature selection in the original data samples, 
$$d_1(\theta) = \mathbb E_{\mathbf X}W[p(Y|\mathbf X),\,p(Y|\mathbf X_{\theta})].$$
It tells us if we select the feature set $\theta$ and how far we deviate from the original conditional distribution. 
Similarly, 
$d_2(\theta)$ means the distance in the noisy labeled data set before feature selection and after feature selection,
$$d_2(\theta) = \mathbb E_{\mathbf X}W[p(\tilde Y|\mathbf X),\,p(\tilde Y|\mathbf X_\theta )].$$

In addition, $\epsilon_2$ 
is defined as the expected Wasserstein distance 
between the conditional distribution in the original clean data sample and the conditional distribution in the noisy labeled data set after feature selection,
$$\epsilon_2 = \mathbb E_{\mathbf X_\theta}W[p(Y|\mathbf X_\theta),\,p(\tilde Y|\mathbf X_{\theta})].$$
Here, the assumption that the same feature set has been chosen in the original data and the noisy labeled data is a bridge to our further analysis. 

Using the above notation, we can rewrite our feature selection optimization Prob.(\ref{origi_opt}) and Prob.(\ref{noise_opt}) simply as:

\begin{align}
\theta_1^* = \underset{|\theta|=K}{\text{argmin}}\quad  d_1(\theta),
\label{theta1} \\
\theta_2^* = \underset{|\theta|=K}{\text{argmin}}\quad d_2(\theta),
\label{theta2}
\end{align}
where $\theta_1^*$ indicates the optimally selected feature set using the original data and $\theta_2^*$ indicates the optimally selected feature set using the noisy labeled data. Although $\theta_2^*$ is the best feature set we can get based on the noisy labeled data set, to give a quantitative evaluation of the optimally selected features based on the noisy labeled data set, we use the quantity 
\begin{equation}
d_1(\theta_2^*) -d_1(\theta_1^*)
\label{feat_qualify}
\end{equation}
as the criterion. Here, $d_1 (\theta_1^*)$ is the baseline and computed based on the clean original data. The smaller of this quantity is, the better of the selected features are. 
In the following, we will show that how this quantity is related to $\epsilon_1$. 

To show the value of $d_1(\theta_2^*) - d_1(\theta_1^*)$, first we demonstrate the relationship between $d_1(\theta)$ and $d_2(\theta)$. Then, we show the relationship between the corresponding optimal selected value of $d_1(\theta_1^*)$ and $d_2(\theta_2^*)$. Finally we will find the value of $d_1(\theta_2^*) - d_1(\theta_1^*)$ is bounded by $4\epsilon_1$. 

Before feature selection, the expected Wasserstein distance between $p(Y|\mathbf X)$ and $p(\tilde Y|\mathbf X)$ is $\epsilon_1$. If the same feature set $\theta$ is selected, will the distance $\epsilon_2$ grow very large? The first conclusion demonstrates this will not happen.

\begin{theorem}
For the same selected feature set $\theta$ in the original data and in the noisy labeled data, $\epsilon_2 \le \epsilon_1$. 
\end{theorem}



\begin{table}[]
\caption{The two optimization problems}
\begin{center}
\begin{tabular}{|c|c|}
\hline
$
\begin{aligned}[t]
\min \quad & \text{tr}(\mathbf Q^\top \mathbf D)\\
\text{s.t.}\quad & \mathbf Q \mathbf 1 = \mathbf p_{Y|\mathbf x}\\
&\mathbf Q^\top \mathbf 1 = \mathbf p_{\tilde Y|\mathbf x} \\
&Q_{i,j} \ge 0, \forall \, i,j 
\end{aligned}
$
& 
$
\begin{aligned}[t]
\min \quad & \text{tr}(\mathbf Q^\top \mathbf D)\\
\text{s.t.}\quad & \mathbf Q \mathbf 1 = \mathbf p_{Y|\mathbf x_\theta}\\
&\mathbf Q^\top \mathbf 1 = \mathbf p_{\tilde Y|\mathbf x_\theta} \\
&Q_{i,j} \ge 0, \forall \, i,j 
\end{aligned}
$ \\
\hline
\end{tabular}
\end{center}
\label{two_opts}
\end{table}

\begin{proof}
In Table \ref{two_opts},  the left optimization problem calculates the distance from the original data to the noisy labeled data; the right optimization problem calculates the distance from the original data to the noisy labeled data after feature selection. To compute $\epsilon_1$ which involves solving the left optimization problem of Table \ref{two_opts} for each configuration of the random vector $\mathbf X = \mathbf x$ and to compute $\epsilon_2$ involves solving the right optimization problem of Table \ref{two_opts}. We denote $\mathbf p_{Y|\mathbf x} = [p(Y=c_1|\mathbf X =\mathbf x),p(Y=c_2|\mathbf X =\mathbf x), \cdots ,p(Y=c_n|\mathbf X = \mathbf x)]^\top$ and the similar vector notation for $\mathbf p_{Y|\mathbf x_\theta}, \mathbf p_{\tilde Y|\mathbf x}$, and $\mathbf p_{\tilde Y|\mathbf x_\theta}$. 

Suppose we have gotten the optimal solution for the left problem as $\mathbf Q(\mathbf x)$, where the optimal solution is denoted as a function of $\mathbf x$. So, the optimal value of the left problem is $\sum_{i,j}Q(\mathbf x)_{i,j}D_{i,j}$. By taking expectation over the random variable $\mathbf X$, we get the expression for the value of $\epsilon_1$ to be $\epsilon_1 = \int f(\mathbf x) \sum_{i,j}Q_{i,j}(\mathbf x)D_{i,j}\,\mathrm{d}\mathbf x$. 

Let another matrix $\mathbf{\tilde Q}(\mathbf x_{\theta})$ with entries defined as 
$$\tilde Q_{i,j}(\mathbf x_\theta) = \frac{\int f(\mathbf x) Q_{i,j}(\mathbf x)\,\mathrm{d}\mathbf x_{\bar \theta }}{f(\mathbf x_\theta)},$$
where $\bar \theta$ is the complement of set $\theta$, i.e. $\bar \theta  = \{1,2,\cdots,M\}\setminus \theta $. In the following, we will show that this matrix satisfies the constraints in the right optimization problem. 

First, we use Bayesian rule to calculate $p(Y=c_i|\mathbf x_\theta)$:
\begin{equation}
p(Y=c_i|\mathbf x_\theta ) = \frac{\int p(Y=c_i|\mathbf x)f(\mathbf x)\,\mathrm{d}\mathbf x_{\bar \theta} }{f(\mathbf x_{\theta })}.
\label{pydx}
\end{equation}

According to the first constraint in the left problem, we have:
$$p(Y=c_i|\mathbf x) = \sum_{j}Q_{i,j}(\mathbf x).$$
Plug this equation into Eq.(\ref{pydx}) and we get
\begin{align*}
p(Y=c_i|\mathbf x_{\theta}) 
&= \frac{\int \sum_{j}Q_{i,j}(\mathbf x)f(\mathbf x)\,\mathrm{d}\mathbf x_{\bar \theta }}{f(\mathbf x_\theta)} \\
&=\sum_j \frac{\int Q_{i,j}(\mathbf x)f(\mathbf x)\,\mathrm{d}\mathbf x_{\bar \theta }}{f(\mathbf x_\theta )} \\
&= \sum_j \tilde Q_{i,j}(\mathbf x_{\theta }).
\end{align*}

It is clear that $\mathbf{\tilde Q}(\mathbf x_{\theta})$ satisfies the first constraint of the right optimization problem. 

Accordingly, following the same routine above, we get
$$p(\tilde Y=c_i|\mathbf x_\theta) = \frac{\int p(\tilde Y=c_i|\mathbf x)f(\mathbf x)\,\mathrm{d}\mathbf x_{\bar \theta }}{f(\mathbf x_\theta)}.$$

Since $p(\tilde Y=c_i|\mathbf x) = \sum_j Q_{j,i}(\mathbf x)$, 
plugging  these into the calculation of $p(\tilde Y=c_i|\mathbf x_\theta)$, we get 
\begin{align*}
p(\tilde Y=c_i|\mathbf x_\theta ) 
&=\frac{\int f(\mathbf x)\sum_j Q_{j,i}(\mathbf x)\,\mathrm{d}\mathbf x_{\bar \theta }}{f(\mathbf x_\theta )} \\
&=\sum_{j}\frac{\int f(\mathbf x) Q_{j,i}(\mathbf x)\,\mathrm{d}\mathbf x_{\bar \theta }}{f(\mathbf x_\theta )} \\
&= \sum_{j}\tilde{Q}_{j,i}(\mathbf x),
\end{align*}
which indicates the joint distribution matrix $\mathbf{\tilde Q}(\mathbf x_\theta )$
also satisfies the second constraint of the right optimization problem. Thus, $\mathbf{\tilde Q}(\mathbf x_\theta)$ is a feasible solution of the right optimization problem. So, 
\begin{align*}
\epsilon_2 
&= \mathbb E_{\mathbf x_\theta }W[p(Y|\mathbf X_\theta )\,,p(\tilde Y|\mathbf X_\theta )] \\
&\le \int \sum_{i,j} \tilde{Q}_{i,j}(\mathbf x_\theta)D_{i,j}f(\mathbf x_\theta )\,\mathrm{d}\mathbf x_\theta \\
&=\int \sum_{i,j}\frac{\int Q_{i,j}(\mathbf x)f(\mathbf x)\,\mathrm{d}\mathbf{x_{\bar \theta }}}{f(\mathbf x_\theta )} f(\mathbf x_\theta )D_{i,j}\,\mathrm{d}\mathbf x_{\theta } \\
&= \int \sum_{i,j}Q_{i,j}(\mathbf x) D_{i,j} f(\mathbf x)\,\mathrm{d}\mathbf x \\
& = \epsilon_1 .
\end{align*}

\end{proof}

It demonstrates that $\epsilon_2$ is always less than or equal to $\epsilon_1$. By constraining selecting the same feature set under the original data and under the noisy labeled data, the value of $\epsilon_2$ is constrained to be no larger than $\epsilon_1$. How about the distance in Eq.(\ref{feat_qualify}) when no equality constraint is put on $\theta_1^*$ and $\theta_2^*$. Theorem 2 tells us that this quantity is also bounded. 

\begin{theorem}
For the optimally selected feature $\theta_1^*$ using the original clean data and optimally selected feature $\theta_2^*$ using the noisy labeled data, the distance 
$d_1(\theta_2^*) - d_1(\theta_1^*)$ is bounded by $4\epsilon_1$. 
That is to say:\\
\begin{equation}
d_1(\theta_2^*) - d_1(\theta_1^*) \le 4\epsilon_1 
\end{equation}.
\end{theorem}

\begin{proof}
Using the triangular inequality property of Wasserstein distance, we get 
\begin{align}
\label{ieq1}
-\epsilon_1-\epsilon_2 \le d_1(\theta_1^*) - d_2(\theta_1^*) \le \epsilon_1+\epsilon_2,  \\ \label{ieq2}
-\epsilon_1-\epsilon_2 \le d_1(\theta_2^*) - d_2(\theta_2^*) \le \epsilon_1+\epsilon_2 ,
\end{align}
where $\theta_1^*$, $\theta_2^*$ have been defined in equation(\ref{theta1})(\ref{theta2}). Since $\theta_1^*$ is the optimal solution for $\min _{|\theta|=K}d_1(\theta)$ and $\theta_2^*$ is the solution for $\min_{|\theta|=K}d_2(\theta)$, we also have 
\begin{align}
d_1(\theta_2^*) \ge d_1(\theta_1^*),
\label{ieq3}\\
d_2(\theta_1^*) \ge d_2(\theta_2^*).
\label{ieq4}
\end{align}
Combine inequalities (\ref{ieq1})(\ref{ieq2})(\ref{ieq3})(\ref{ieq4}) and we have
\begin{align}
d_1(\theta_2^*) - d_1(\theta_1^*) + d_2(\theta_1^*) - d_2(\theta_2^*) \le 2\epsilon_1+2\epsilon_2.
\end{align}
Since $d_1(\theta_2^*) -d_1(\theta_1^*)\ge 0$, $d_2(\theta_1^*)-d_1(\theta_2^*)\ge 0$ and $\epsilon_2 \le \epsilon_1$, we have 
\begin{equation}
d_1(\theta_2^*) - d_1(\theta_1^*) \le 4 \epsilon_1 
\label{results}.
\end{equation}
\end{proof}

Theorem 2 shows that the optimal feature set $\theta_2^*$ under noisy labeled data is $4\epsilon_1 $ suboptimal in the original clean data. 

However, we should notice that the value $\epsilon_1$ is fixed for a given data set and it is determined by the physical system.
Although we can not change this value, we wish this value to be small. If this value is large, which means the data set quality is bad, it can not provide us enough information to select useful feature set. 
If we can get an estimated value of $\epsilon_1$ and 
get further analysis, 
it would be greatly helpful. In the following, we will give some insights into the value of $\epsilon_1$ and show this value will not be too large. 

To compute $\epsilon_1$, we need to solve the optimization problem in the left of Table \ref{two_opts} for every configuration of $\mathbf X = \mathbf x$. 
Because this optimization problem does not have a closed form solution in general cases, we can not get a closed form solution for $\epsilon_1$. Also, the relationship between $p(Y|\mathbf X)$ and $p(\tilde Y|\mathbf X)$ is not specified, which makes this problem even harder to solve. Instead of directly solving this problem, we can find some meaningful bound for $\epsilon_1$. 


There are two most widely used noisy label models, which are \emph{noise at random} (NAR) and \emph{noise not at random} (NNAR)\cite{frenay_classification_2014}. 
NAR model makes assumption that the observed label depends on the underlying true label. So, the observed class conditional distribution given the feature $\mathbf X$
is:
\begin{equation}
p(\tilde Y|\mathbf X) = \sum_{Y} p(\tilde Y|Y)p(Y|\mathbf X)
\label{nar-model},
\end{equation}
where $p(\tilde Y|\mathbf Y)$ denotes the transition probability which quantifies the the probability that the true label is $Y$ while observed is $\tilde Y$. Under this model, it is still hard to get closed form solution of $\epsilon_1$. However, we can construct a joint distribution matrix and get it's upper bound. 
Let the elements of joint distribution matrix be $Q_{i,j}(\mathbf x) = p(Y=c_i,\tilde Y =c_j|\mathbf x)$ and $p(Y=c_i,\tilde Y =c_j|\mathbf x) = p(\tilde Y=c_j|Y=c_i)p(Y=c_i|\mathbf x)$. We can verify that this matrix is a feasible solution of the left problem of Table \ref{two_opts}. So, we have 
\begin{align}
\epsilon_1 \nonumber 
&\le \int p(\mathbf x) \sum_{i,j} Q_{i,j}(\mathbf x)D_{i,j} ,\mathrm{d}\mathbf x  \\ \nonumber 
& = \int f(\mathbf x) \sum_{i,j} p(Y=c_i,\tilde Y=c_j|\mathbf x)D_{i,j} \mathrm{d}\mathbf x \\ 
& = \sum_{i,j} p(Y=c_i,\tilde Y=c_j) D_{i,j},
\end{align}
where $p(Y=c_i,\tilde Y=c_j)$ indicates the joint probability that the true label is $c_i$ while the observed is $c_j$. If two classes, $c_i,c_j$, are similar, we should expected that the transition probability $p(\tilde Y=c_j| Y=c_i)$ should be large and so is the joint probability $p(Y=c_i,\tilde Y=c_j)$. 

The NNAR model assumes that the observed label depends on both the true label and its features. So, the observed class conditional distribution given the features is
\begin{equation}
p(\tilde Y|\mathbf X) = \sum_{Y} p(\tilde Y|Y,\mathbf X)p(Y|\mathbf X).
\label{nnar-model}
\end{equation}
Compared with Eq.(\ref{nar-model}), it only differs in the form of the transition probability. Construct a feasible joint probability matrix as $Q_{i,j}(\mathbf x) = p(Y =c_i,\tilde Y = c_j|\mathbf x)$ and $p(Y=c_i,\tilde Y=c_j|\mathbf x) = p(\tilde Y =c_j |\mathbf Y=c_i,\mathbf x) p(Y=c_i|\mathbf x)$. Under this noise model, we can also find an upper bound of $\epsilon_1$ as
\begin{align}
\epsilon_1  \nonumber 
& \le \int f(\mathbf x) \sum_{i,j} p(Y=c_i,Y=c_j|\mathbf x),\mathrm{d} \mathbf x \\
&=\sum_{i,j}p(Y=c_i,\tilde Y=c_j)D_{i,j}.
\end{align}
This upper bound has the exact same form as the upper bound induced by NAR model except the way of computing the joint distribution.    

Generally, let $p(Y,\tilde Y|\mathbf X)$ denotes the joint conditional distribution of $Y$ and $\tilde Y$ given feature $\mathbf X$, where $Y$ is the original label  and $\tilde Y$ is the noisy label 
and it satisfies the constraints of the optimization problem, that is, its marginal conditional distributions are $p(Y|\mathbf X)$ and $p(\tilde Y|\mathbf X)$ respectively. Then we get an upper bound for optimal value of the optimization problem 
$$\sum_{i,j} p(Y=c_i,\tilde Y= c_j|\mathbf X=\mathbf x) D_{i,j}.$$
By taking the expected value regarding $\mathbf X$, we get the upper bounded solution of $\epsilon_1$ 
\begin{align}
\nonumber 
\epsilon_1 
& \le \int f(\mathbf X = \mathbf x) \sum_{i,j} p(Y=c_i,\tilde Y=c_j|\mathbf X=\mathbf x)D_{i,j}\, \mathrm{d}\mathbf x  \\ \nonumber 
&= \sum_{i,j} D_{i,j} \int f(\mathbf X=\mathbf x) p(Y = c_i,\tilde Y=c_j|\mathbf X= \mathbf x)  \, \mathrm{d}\mathbf x \\
& = \sum_{i,j}p(Y=c_i,\tilde Y=c_j)D_{i,j}.
\label{bound_e1}
\end{align}
Different from other algorithms whose bound analysis are based on specific noise label model, this bound only depends on the distance matrix and the joint distribution of original label and noisy label.

To evaluate the bound, we need look inside the bound, the term $p(Y=c_i,\tilde Y = c_j) D_{i,j}$. This term involves the joint probability of the original label and noisy label and the distance matrix. 
If we admit that the larger distance between two classes is, the lower of their joint probability, that is to say, if the distance is large, the joint probability $p(Y=c_i,\tilde Y = c_j)$ should be small, and vice versa, the term would not be too large. 
Other information measure based algorithms, like the KL-distance-based algorithm, will result in exploding value of the distance from original to noisy labeled. For example, when calculate the KL distance of the two conditional probabilities, $\sum_i p(Y=c_i|\mathbf x) \frac{p(Y=c_i|\mathbf x)}{p(\tilde Y=c_i|\mathbf x)}$ and if the noisy label lead to $p(\tilde Y=c_i|\mathbf x)$ approximate to zero, we would get the  KL distance exploded. Then the algorithm based on KL distance will not provide useful information for how far the noisy labeled data deviates from the original data.

\section{Simulation}
In this section, we will do three experiments to demonstrate the superior of our algorithm.

In the first experiment, we use the Enron email corpus data set\cite{klimt2004enron}, and some pre-processing  had been done in \cite{kocev2012ensembles}. The data set contains 1648 samples. Each sample includes a 1001 dimensional feature vector $\mathbf x_i$ and multiple labels $\mathbf y_i$. Each dimension of the feature indicates one word in the email. Based on the preprocessing in \cite{kocev2012ensembles}, if a word appears in the email, then its corresponding feature will set to be one. So, each  feature is binary valued. The labels of the samples indicate the folders where the email belongs to. For example, the label "inbox/primary/discussion" means the email in the inbox folder and primary sub-folder and discussion sub-sub-folder. So, the label itself has a hierarchic structure. 

\begin{figure}[t]
\centering
\includegraphics[width=.3\linewidth]{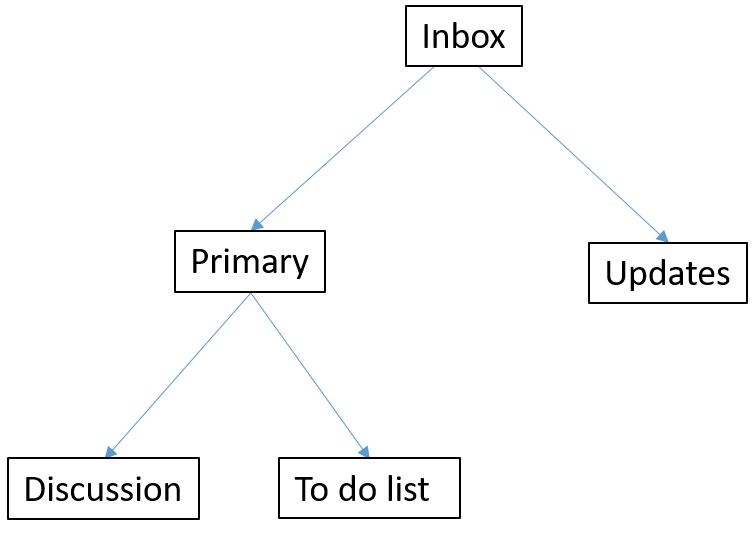}
\includegraphics[width=.3\linewidth]{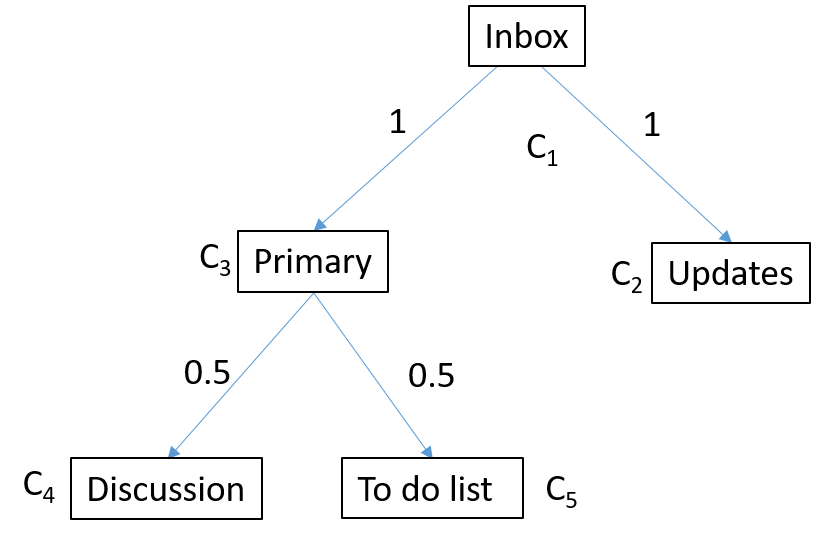}
\includegraphics[width=.3\linewidth]{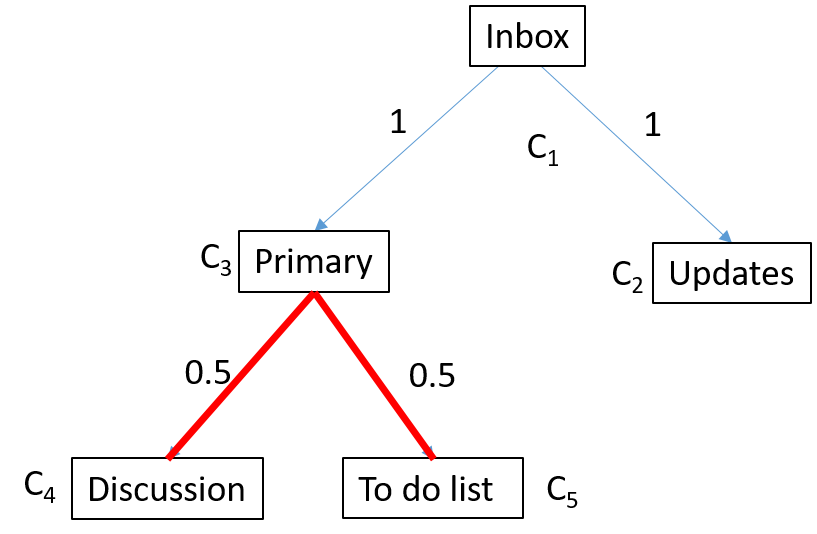}
\caption{The three pictures from the left to the right show the hierarchic structure of the labels, the weight assigned to the path between hierarchic labels, and one example to compute the class distance.}
\label{labelstruct}
\end{figure}

The picture on the left of Figure(\ref{labelstruct}) shows one example of the label structure. In this picture, the root node is the label "inbox". However, there is no root node in the Enron dataset. In the experiment, we virtually assign a root node in this hierarchic structure. 
Then, the label structure can be depicted as a tree. The nodes of the tree indicate the class labels. One important thing to be notified is that not only the leaves of the tree represent labels, but also the non-leaf nodes can represent labels. With this tree structure, it provides us a natural way to compute the distance between different classes. The distance between class $i$ and class $j$ is denoted as $D_{ij}=d(c_i,c_j)$. To compute the distance between classes in our case, we first assign weights to the paths between classes. Here we assume the nearer to the root is, the larger weights should be assigned to the paths. 
The philosophy behind this is that the higher the layer of the nodes is, the larger the disparity of the classes is. 
Then we can define the class distance, $d(c_i,c_j)$, as the shortest path from class $i$ to class $j$. The second picture shows one possible way to assign the weights to the paths. The third picture in Fig.\ref{labelstruct} shows how to compute the distance between class $4$ and class $5$. Clearly, under this definition, the distance satisfies the triangular inequality property, i.e. $d(c_i,c_j)\le d(c_i,c_k) + d(c_k,c_j)$. 

The Enron dataset labels have a three-layer structure. From the root layer to the third layer, we assign the weights of the paths to be $0.5,0.2$, and $0.05$ respectively. Even though Enron data is a multi-label dataset, we treat it as single labeled data set by doing the following pre-processing.
For each sample  with multiple labels in the training dataset, we assign the same features to each individual  labels and treat it as an independent sample.  Using the feature selection algorithm on the training dataset, we can get our selected feature set index $\theta$. Then based on the selected feature set, we retain the features of the training and testing indicated by the feature set. Using the reduced training data, we train a classifier and use this classifier to predict the labels for the testing data set. Here, in our experiment, we use the k nearest neighbor (kNN) classifier. We use the kNN classifier because the classifier makes few assumptions on the data set and has no hyper-parameters needing to be tuned. 

To evaluate the results, we use the top-k loss. It is defined as:
$$loss = \frac{1}{k}\sum_{i=1}^k \min_{y\in \mathcal{L}}d(\hat y_i, y).$$
where $\mathcal{L}$ is the true labels and $\hat y_i$ is the predicted label with the $i^{th}$ highest prediction probability. 
This top-k loss is more general than the classification error. It does not only assign $0$ to a accurate prediction and $1$ to wrongly prediction but also how far the wrongly predicted label are to the ground truth. 

In this experiment and the following experiments, we set the number of conditional features $L$ to be $3$.   

\begin{figure}[h]
\centering
\includegraphics[width=.8\linewidth]{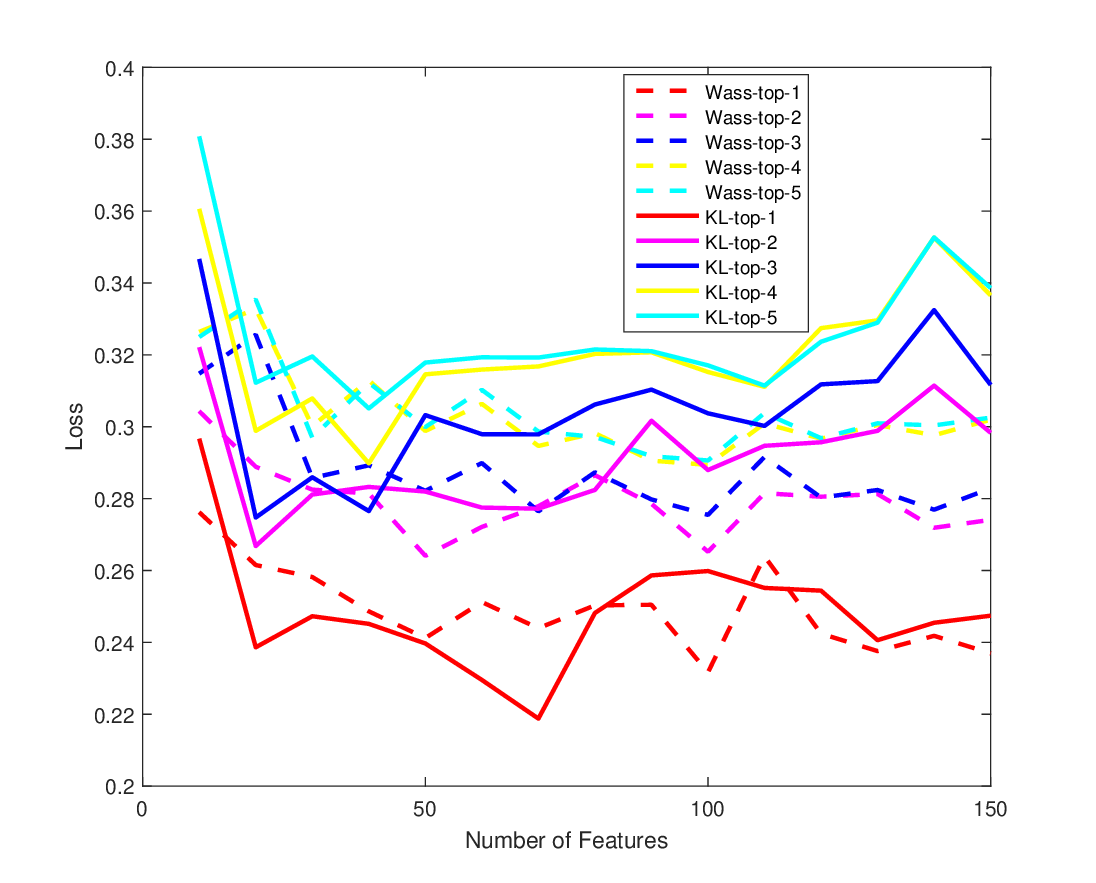}
\caption{The top-k loss with different k and different number of features.}
\label{topkloss}
\end{figure}

Figure(\ref{topkloss}) shows the top-k loss with different k and different numbers of features. Our algorithm is denoted as 'Wass' while the KL-distance-based algorithm is denoted as 'KL'. From the figure we can see, our Wasserstein -distance-based algorithm has smaller top-k loss in most cases, especially when k is large. 
And the top-k loss of the two algorithms both increase as k increase.
\begin{figure}[h]
\centering
\includegraphics[width=.45\linewidth]{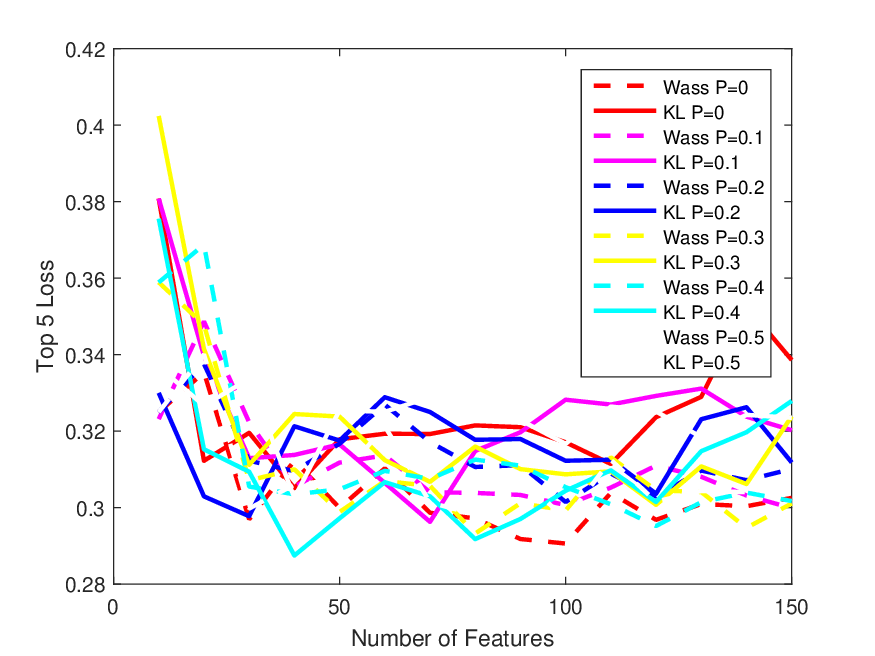}
\includegraphics[width=.45\linewidth]{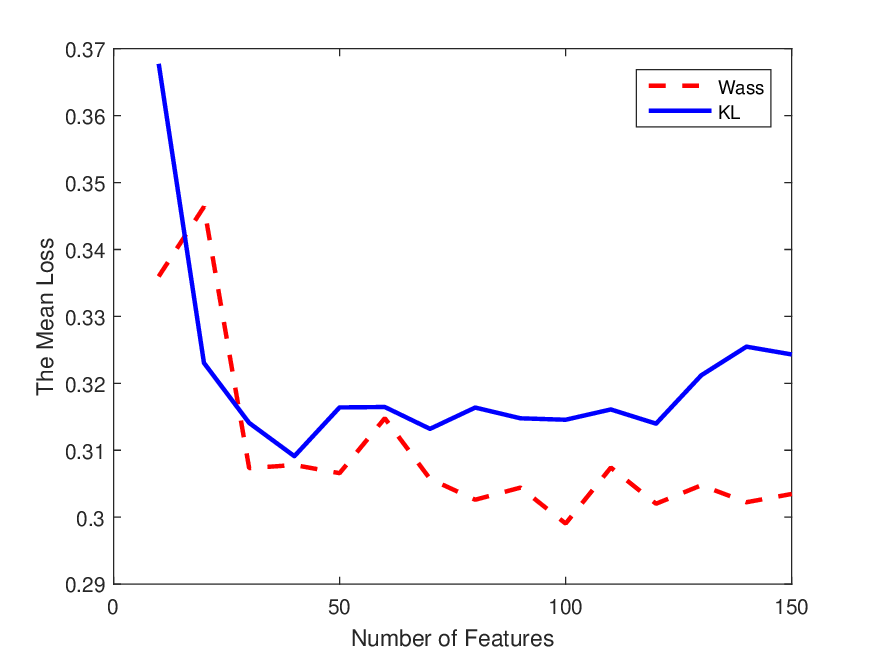}
\caption{The left picture shows top-5 loss with different level of label flipping noises and different numbers of features in two algorithms; the right picture shows the averaged loss over different flipping noises described in the left picture.}
\label{noisyloss}
\end{figure}

In second simulation, we will show the robustness of our algorithm under noisy labels. In this experiment, we randomly flip the labels in the training data set to one of its nearest neighbors with probability $P$, where the nearest neighbors are defined as the labels which have the class distance less than or equal to 0.2. We use feature selection algorithms on this noisy labeled training data and get the selected feature set. Like in the previous experiment, we reduce the feature dimension for the training and testing dataset. Then we use the feature-reduced training data with the right labels to train a classifier, and then we use the classifier to predict the testing data labels. 

Figure(\ref{noisyloss}) demonstrates the top-5 losses under different noise levels and different numbers of features using these two algorithms. The left picture shows our algorithm has smaller top-5 losses and behave much more stably than the KL-distance-based algorithm. Then, to make the left picture more clear, we average the top-5 losses over different $P$. This picture better shows our algorithm has lower top-5 losses. For KL distance algorithm, the behavior is a little pathologic. 
With the number of features increase, the KL distance based algorithm's top-5 losses even increase when the number of features is over 120.
This is due to the fact that under the noisy setting, the KL-distance-based algorithm can not learn the most useful features.

\begin{figure}[h]
\centering
\includegraphics[width=.8\linewidth]{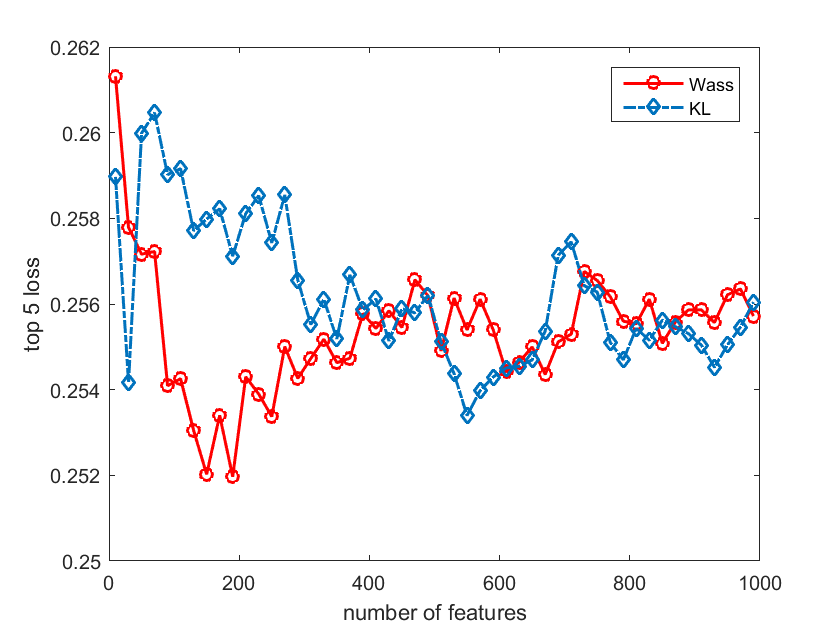}
\caption{top-5 loss with different number of features on the Flickr data set.}
\label{flickrresults}
\end{figure}

In the second experiment, we use the Yahoo/Flickr Creative Commons 100M dataset\cite{thomee2015new}\cite{frogner_learning_2015}. It is a multi-label dataset. Each sample is an image with some tags given by the users. We do some pre-processings similarly to \cite{frogner_learning_2015}. Here, we give a summary of this pre-processing. We randomly choose 10,000 samples as the training dataset and another 10,000 samples as the testing dataset. The features we used are extracted using the pre-trained neuron network, MatConvNet\cite{vedaldi2015matconvnet} and there are totally 4096 features. We discretize each feature into 10 discrete numbers based on their quantile value. We extract all the tags and use word2vector\cite{mikolov2013distributed} technology to map each tag into a low dimensional unit vector. Then, we use the Euclidean distance between those vectors as the distance between tags, and carry out the feature selection on the training dataset. 
The classifier we use is kNN and the loss function is the top-k cost as the same in the previous experiment. Fig.\ref{flickrresults} shows the top-5 loss under different numbers of features. From the figure we can see, when the number of features is relatively small, say below 10\%, our Wasserstein-distance-based algorithm has less top-5 loss than the KL-distance-based algorithm. Which indicates our algorithm has the power to learn more important features. When the number of features continues increasing, the loss for both of the algorithms increase. 
It is due to overfitting considering the number of features is large and the training samples is not so many.

\begin{table}[t]
\centering
\caption{20 newsgroups dataset topics}
\begin{tabular}{|l|l|l|}
\hline
\begin{tabular}[c]{@{}l@{}}comp.graphics\\ comp.os.ms-windows.misc\\ comp.sys.ibm.pc.hardware\\ comp.sys.mac.hardware\\ comp.windows.x\end{tabular} & \begin{tabular}[c]{@{}l@{}}rec.autosrec.motorcycles\\ rec.sport.baseball\\ rec.sport.hockey\end{tabular} & \begin{tabular}[c]{@{}l@{}}sci.crypt\\ sci.electronics\\ sci.med\\ sci.space\end{tabular}         \\ \hline
misc.forsale                                                                                                                                        & \begin{tabular}[c]{@{}l@{}}talk.politics.misc\\ talk.politics.guns\\ talk.politics.mideast\end{tabular}  & \begin{tabular}[c]{@{}l@{}}talk.religion.misc\\ alt.atheism\\ soc.religion.christian\end{tabular} \\ \hline
\end{tabular}
\label{20newsgroupstopics}
\end{table}

\begin{figure}[h]
\centering
\includegraphics[width=0.8\linewidth]{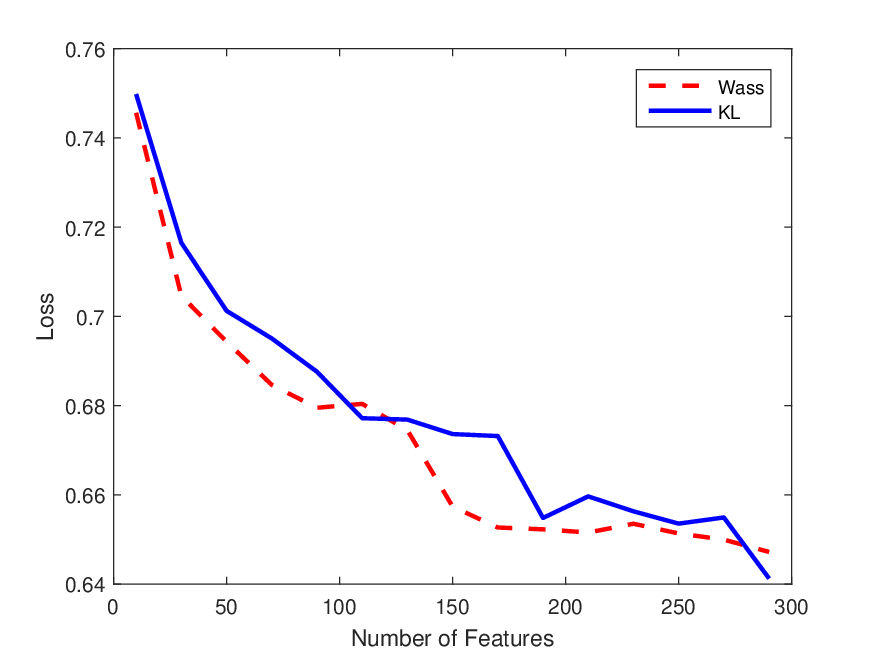}
\caption{The result using the 20 newsgroups dataset.}
\label{20news}
\end{figure}

In the third experiment, we use the 20 newsgroups dataset\cite{joachims1996probabilistic}. It is a collection of 20,000 documents across 20 different newsgroups. The topics of the 20 newsgroups are summarized in Table \ref{20newsgroupstopics}. In this table, the topics are partitioned according to subject matter. Within each individual cell the topics are very closely related, while between two different cells the topics are unrelated. With the advantage of topic diversity, we set the class distance to be $0.2$ within each topic cell and class distance to be $1$ between different topic cells. The features of the dataset is the top 20000 vocabularies ordered by the term frequency. The value of each feature is computed by uniformly discretizing its tf-idf value into 10 digital numbers. Then we do feature selection with two different algorithms after this pre-processing. 

Fig.\ref{20news} shows the results of two different algorithms. Since the dataset is single labeled, we use the top-1 cost as the loss function and the classifier is kNN as the previous experiments. As we can see, our algorithm provides smaller loss at most cases.

\section{summary and future work}
We introduce the Wasserstein distance into one of the traditional feature selection frameworks and demonstrate its potentials in feature selection. 

The Wasserstein distance based feature has a promising prospect. However, the distance matrix used in the algorithm is hard to define and it varies from datasets to datasets. One future work is to find a unified way to calculate the class distance. 

\ifCLASSOPTIONcaptionsoff
  \newpage
\fi

\bibliography{mybib.bib}
\bibliographystyle{IEEEtran}
\end{document}